\documentclass[10pt,twocolumn,letterpaper]{article}

\usepackage{wacv}
\usepackage{times}
\usepackage{epsfig}
\usepackage{graphicx}
\usepackage{amsmath}
\usepackage{amssymb}
\usepackage{booktabs}
\usepackage{array}
\usepackage{multirow}
\usepackage{breakcites}
\usepackage{subfig}

\newcommand{\modelname}{TransPillars}

\def\wacvPaperID{118} 

\wacvalgorithmstrack   

\wacvfinalcopy 

\ifwacvfinal
\usepackage[breaklinks=true,bookmarks=false]{hyperref}
\else
\usepackage[pagebackref=true,breaklinks=true,colorlinks,bookmarks=false]{hyperref}
\fi

\begin{document}

\title{TransPillars: Coarse-to-Fine Aggregation for Multi-Frame 3D Object Detection}

\author{
Zhipeng Luo$^{1,3}$ \;
Gongjie Zhang$^{1}$ \;
Changqing Zhou$^{1,3}$ \; \\
Tianrui Liu$^{1,3}$ \; 
Shijian Lu$^1$\thanks{Corresponding author} \; 
Liang Pan$^{2}$ \;\\
$^1$ Nanyang Technological University 
$^2$ S-Lab, Nanyang Technological University 
$^3$ Sensetime Research \\
}


\maketitle
\thispagestyle{empty}

\begin{abstract}
3D object detection using point clouds has attracted increasing attention due to its wide applications in autonomous driving and robotics. However, most existing studies focus on single point cloud frames without harnessing the temporal information in point cloud sequences. In this paper, we design \modelname{}, a novel transformer-based feature aggregation technique that exploits temporal features of consecutive point cloud frames for multi-frame 3D object detection. \modelname{} aggregates spatial-temporal point cloud features from two perspectives. First, it fuses voxel-level features directly from multi-frame feature maps instead of pooled instance features to preserve instance details with contextual information that are essential to accurate object localization. Second, it introduces a hierarchical coarse-to-fine strategy to fuse multi-scale features progressively to effectively capture the motion of moving objects and guide the aggregation of fine features. Besides, a variant of deformable transformer is introduced to improve the effectiveness of cross-frame feature matching. Extensive experiments show that our proposed \modelname{} achieves state-of-art performance as compared to existing multi-frame detection approaches. Code will be released.
\end{abstract}

\begin{figure}
    \centering
    \includegraphics[width=1.0\linewidth]{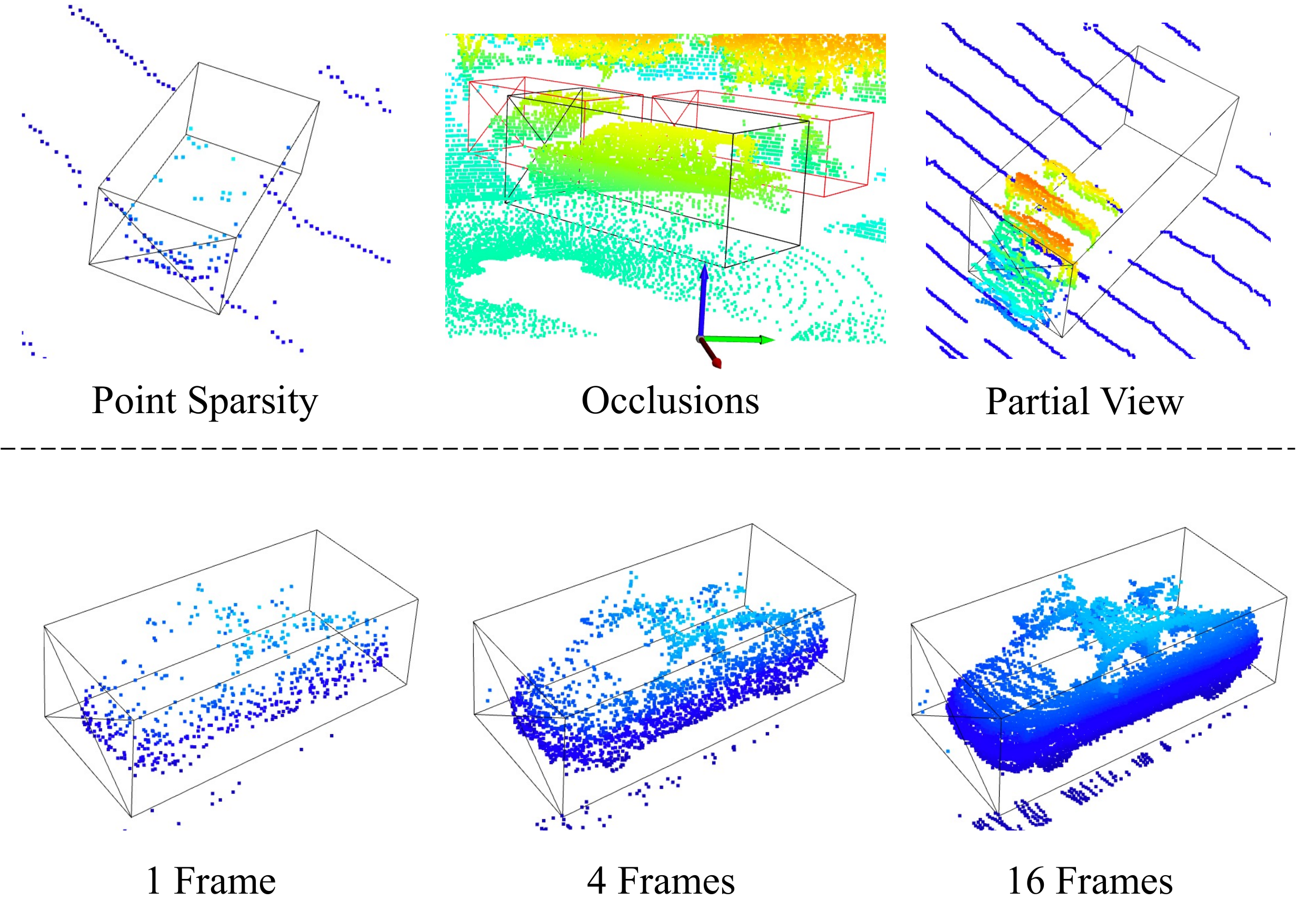}
    \vspace{-4.0mm}
    \caption{
    Upper: Illustration of challenges in point cloud object detection. Occluded objects are indicated by red boxes and LiDAR is highlighted by the axes. Lower: Sequential frames contain complementary information and aggregating the complementary information leads to a more complete view.
    }
    \label{fig:teaser}
    \vspace{-3mm}
\end{figure}

\section{Introduction}

3D object detection using point clouds has been actively studied in recent years due to its wide applications in the fields of autonomous driving and robotics. With the advent of large-scale datasets \cite{geiger2013kittidataset, sun2020waymo, caesar2020nuscenes}, popular deep learning-based single-frame 3D object detectors \cite{zhou2018voxelnet, shi2019pointrcnn, shi2020pvrcnn, yin2021centerpoint, qi2019votenet, yang2019std, yang20203dssd} are proposed. However, there are challenges difficult to resolve with a single sweep of point cloud. \textit{First}, point clouds are sparse, especially at a long distance away from the LiDAR sensor. \textit{Second}, incomplete point clouds due to partial observation, occlusion, and view truncation lead to ambiguities in the object geometry (as shown in Fig.~\ref{fig:teaser} upper). 

Leveraging multiple frames of point clouds, on the other hand, provides critical temporal cues that mitigate the above-mentioned challenges \cite{yang20213dman, qi2021offboard, yu2020leveraging}. The lower section of Fig.~\ref{fig:teaser} illustrates that points accumulated from consecutive frames progressively form a holistic depiction of the object. Nevertheless, leveraging temporal information is not trivial. A straightforward approach is simply concatenating points from multiple frames \cite{caesar2020nuscenes}. However, such a method does not explicitly model cross-frame relations and the performance deteriorates for moving objects as the number of concatenated frames exceeds a certain threshold \cite{yang20213dman}. Some recent approaches resort to fusing multi-frame information via feature aggregation \cite{luo2018fast, huang2020lstm}. In particular, inspired by relational networks \cite{hu2018relation}, \cite{yang20213dman} proposes a 3D multi-frame attention network that performs feature alignment and aggregation on pooled instance-level features. Despite its remarkable performance gain as compared to the single frame baseline, the RoI pooling process inevitably leads to 1) loss of instance details due to the misalignment between RoI and the object as dimension and location estimation is imperfect, 2) loss of contextual information due to separation of the objects from the scene, both undermining cross-frame correlation modeling. The proposed feature pooling method also introduces extra class-specific heuristic designs, such as the number of key points for each class, which make the method less adaptable. Moreover, such a method depends on high-quality region proposals, which are not guaranteed under the above challenges associated with point clouds.

Motivated by the above observations, in this paper, we explore aggregating multi-frame information directly from feature maps. Specifically, we propose \modelname{}, which builds on top of PointPillars \cite{lang2019pointpillars} and employs the attention mechanism of transformer to perform cross-frame feature aggregation at voxel-level. To avoid prohibitive computational complexity and memory consumption, instead of performing global attention computations between each pair of tokens from the feature maps as in regular transformers, we adopt deformable attention \cite{zhu2020deformable} for feature aggregation that each voxel adaptively attends to a small number of target voxels. To address the limitations of the original deformable attention in the cross-frame matching process, we develop a variant of deformable attention by incorporating the query-key matching operation in the attention module to better adapt to moving objects. In order to effectively capture the motion of fast-moving objects, we introduce a novel coarse-to-fine aggregation strategy to first identify the high-level instance correspondences to guide the subsequent fusion of fine features for accurate localization.

In summary, the contributions of this work are threefold. \textit{First}, We propose a transformer-based multi-frame point cloud detection model named \modelname{}. \modelname{} performs voxel-level feature aggregation instead of using pooled instance features to preserve instance details and exploit contextual information for accurate localization. We show that the performance of our proposed method surpasses state-of-the-art multi-frame approaches on standard benchmarks. \textit{Second}, we design a novel hierarchical coarse-to-fine feature aggregation strategy to effectively capture the motion of fast-moving objects to guide subsequent fusion of fine features. \textit{Third}, we develop a variant of deformable attention for effective cross-frame feature matching in the feature aggregation process.

\section{Related Work}

\noindent\textbf{Single-frame Point Cloud Object Detection.\;\;\;} Single-frame 3D detectors can generally be classified into two categories, namely voxel-based methods and point-based methods. Voxel-based methods \cite{zhou2018voxelnet, kuang2020voxel, yan2018second, shi2020parta2, yin2021centerpoint} first project points to grids of fixed size to form voxel representations and process the input with convolutions. In particular, 3D sparse convolution \cite{graham20183dsparseconv} is widely used to speed up the computation by leveraging the sparsity of point cloud and only considering the non-empty voxels. On the other hand, point-based approaches \cite{qi2018frustum, shi2019pointrcnn, qi2019votenet, yang2019std, yang20203dssd} first utilize point cloud feature extractors \cite{qi2017pointnet, qi2017pointnet++} to perform point sampling and feature extraction for subsequent region proposal generation or bounding box prediction. Recent method \cite{shi2020pvrcnn} propose to combine voxel-based and point-based approaches to obtain refined predictions. In this work, we adopt PointPillars \cite{lang2019pointpillars}, a lightweight single stage voxel-based detector commonly used by multi-frame detectors \cite{yin2020lidarconvgru, yang20213dman, yuan2021temporalTCTR}, as our base model.

\noindent\textbf{Multi-frame Point Cloud Object Detection.\;\;\;} A basic approach to utilize multiple frames for point cloud detection is through point concatenation of consecutive frames \cite{caesar2020nuscenes}. Despite its effectiveness, the performance gain over the single-frame method is limited especially over longer time intervals since it does explicitly model the relations among frames \cite{yang20213dman}. FaF \cite{luo2018fast} proposes to concatenate feature maps extracted from point clouds instead, but faces the similar limitation of misaligned representations. Some recent approaches \cite{huang2020lstm, yin2020lidarconvgru} employ recurrent networks to aggregate multi-frame features but such methods often lead to high computational cost. \cite{qi2021offboard} explores an offboard setting for auto-labeling by performing detection for individual frames and aggregating the results from the entire sequence. Inspired by relational networks \cite{hu2018relation} and its applications \cite{chen2020memory, shvets2019leveraging, wu2019sequence, geng2020object, han2020mining} on 2D video object detection, 3D-MAN \cite{yang20213dman} proposes to apply attention mechanisms on pooled RoI features for multi-frame alignment and aggregation. However, RoI pooling separates the objects from the context and leads to the loss of details. Our method also resorts to attention mechanisms but performs feature aggregation at voxel-level to maximally preserve the instance details and contextual information.

\begin{figure*}[t]
    \centering
    \includegraphics[width=0.95\linewidth]{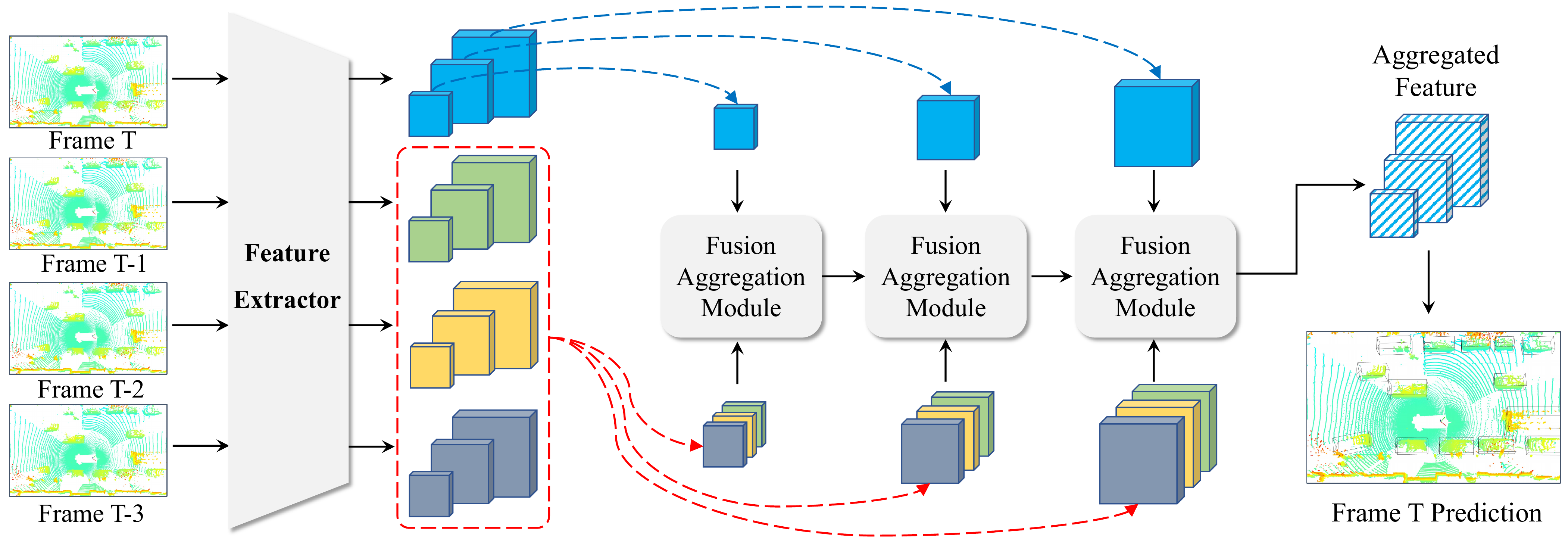}
    \vspace{-3mm}
    \caption{
    The framework of the proposed \modelname{}: Given multiple consecutive point cloud frames as inputs, multi-scale features are first extracted by a feature extractor which are then aggregated with multiple Fusion Aggregation Modules in a coarse-to-fine manner. We first aggregate coarse features to extract high-level cross-frame correspondences and then employ them to guide the aggregation of fine features. Finally, the multi-scale aggregated features are fused to produce the final predictions. Best viewed in color. 
    }
    \label{fig:model}
\vspace{-3.0mm}
\end{figure*}

\noindent\textbf{Vision Transformers.\;\;\;} Transformers~\cite{vaswani2017attention} were first proposed in natural language processing (NLP) as an attention-based building block, which allows for information aggregation from the entire input sequence. In recent years, transformers have been adopted to tackle various computer vision problems and achieved remarkable success \cite{dosovitskiy2020imagevit, carion2020detr, dosovitskiy2020imagevit, wang2021pyramid, zheng2021rethinking, hudson2021generative, zhang2021meta, zhang2022accelerating, zhou2022pttr}. One main advantage of transformers is their global receptive field and capability in capturing long-range relationships. However, such characteristics also bring unfavorable high computational cost and memory usage, which makes it challenging for applications that involve large-scale inputs or feature representations. Extensive studies have been carried out to address this issue and a number of researches \cite{liu2021swin, dong2021cswin, huang2021shuffle, chu2021twins, fang2021msg, yang2021focal} propose to reduce the complexity with restricted attention patterns such as local windows. Such approaches may not be suitable for our task as objects are moving at various speeds, which makes it difficult to determine a suitable window size. On the other hand, \cite{zhu2020deformable} proposes deformable attention, which learns query-dependent sparse sampling locations to adaptively gather features from the value input. However, deformable attention does not explicitly enforce query-key matching, which leads to limited capability in the cross-frame feature matching in our task. In this work, we propose a variant of deformable attention to mitigate this issue.

\section{Method}
In this section, we introduce our proposed \modelname{} for multi-frame point cloud object detection in detail. Section~\ref{subsec:preliminary} briefly introduces the PointPillars~\cite{lang2019pointpillars} model, which is used as the base model in our proposed method. Section~\ref{subsec:overview} gives an overview of our proposed method, followed by Sections~\ref{subsec:FAM} and \ref{subsec:attention}, which explain the Fusion Aggregation Module and our proposed attention mechanism. Lastly, we describe the losses for model optimization in Section~\ref{subsec:losses}.

\subsection{Preliminaries}
\label{subsec:preliminary}
We use PointPillars as the base model in our proposed method. PointPillars differs from regular voxel-based detectors that it only discretizes the input point cloud with a grid of fixed size in the x-y plane, thus forming pillars instead of cubic voxels. The points inside each pillar are then augmented and used to generate a feature vector. Note that only non-empty pillars are processed by the network to speed up the feature extraction. The obtained pillar features are then scattered back to their corresponding locations in the scene to form a pseudo-image representation. Subsequently, the pseudo-image features are processed by a feature pyramid network, which uses convolution layers to extract multi-scale features ${\mathbf{F}}^i$, where $i\in\{1,2,3\}$ denotes the scale level that $i=1$ refers to the feature map of the smallest scale. Finally, the downsampled features are upsampled with transpose convolutions, and the final prediction is generated using a detection head based on the concatenated features. We refer the readers to \cite{lang2019pointpillars} for details. 

PointPillars is known as a lightweight 3D detector that achieves a good balance between efficiency and accuracy, and it has been widely adopted by multi-frame 3D detection approaches \cite{yang20213dman, yin2020lidarconvgru, yuan2021temporalTCTR} as the baseline model. We also base our proposed method on PointPillars without making modifications to the baseline model and focus on the multi-frame feature aggregation.

\subsection{\modelname{}}
\label{subsec:overview}
Fig.\ref{fig:model} gives an overview of our proposed \modelname{}. In the feature extraction stage, given a sequence of point cloud input $\{I_{T-n}\}_{n=0}^{N-1}$, where $N$ denotes the number of frames in the sequence and $I_T$ refers the current (latest) frame, a shared feature extractor of the base model is used to extract multi-scale features $\{\mathbf{F}_{T-n}\}_{n=0}^{N-1}$. In the subsequent feature aggregation stage, the goal is to aggregate useful information from the past frames to enrich the feature representations of the current frame in order to enhance the detection prediction. Therefore, we regroup the extracted features into current and past representations and perform attention-based feature aggregation.

As introduced in the related works, a commonly adopted paradigm in the 2D video detection field is to extract instance-level features with RoI Pooling or RoIAlign operations and perform feature fusion with relation networks \cite{hu2018relation} or its variants. However, unlike image frames which are essentially projections from the 3D real-world space to the flattened 2D space, point clouds possess different characteristics. First, due to the loss of depth information in projection, images do not contain explicit spatial information such as object size and location, while point clouds offer accurate 3D coordinates. Moreover, as compared to typical video detection datasets such as ImageNet VID \cite{russakovsky2015imagenet}, object movements pose a bigger challenge in point cloud applications such as autonomous driving where rapid moving objects are often observed. Region pooling methods separate objects from the scene and lead to loss of instance details as well as contextual information, which restricts the effectiveness of feature aggregation. Motivated by the above analysis, we propose to directly perform voxel-level aggregation from feature maps to maximally exploit the context and preserve details for accurate localization. 

To perform voxel-level aggregation, two challenges are faced. The first challenge is the high computational cost and memory consumption associated with the attention mechanism due to the global matching process, which can be prohibitive for large-scale point cloud scenes studied in this work. To alleviate this issue, we resort to the recently proposed deformable attention \cite{zhu2020deformable}, where each element in the query only adaptively attends to a small number of value tokens. However, the original deformable attention does not explicitly enforce query-key matching, which makes it less effective for our multi-frame detection task where cross-frame feature matching is required. Therefore, we introduce a variant of deformable attention that incorporates the said matching process, which will be detailed in Section \ref{subsec:attention}. In practice, in order to further reduce the computation, we select a proportion of the voxels from the current frame with high classification scores predicted by the base model to form the query feature.

The second challenge is to capture the motion of fast-moving objects to establish the cross-frame correspondence. To achieve this goal directly on fine features is difficult due to the large searching space brought by the small voxel size.
To address this issue, we design a novel coarse-to-fine aggregation strategy by leveraging the multi-scale features generated by the base model. Specifically, we employ a transformer-based Fusion Aggregation Module (FAM) to perform feature fusion and feature aggregation starting from the coarse feature maps. The output aggregated features are then fused with the feature maps of the next scale level for subsequent aggregations. The main idea is to utilize coarse feature maps to perform a coarse matching for cross-frame feature tokens, and the aggregated features are used to guide the following matching process of fine features whose tokens are of smaller physical size. Finally, the outputs of all FAMs are combined as in PointPillars \cite{lang2019pointpillars} to generate the final prediction. The details of FAM are elaborated in Section \ref{subsec:FAM}.

We highlight that our proposed method emphasizes feature aggregation and does not require modifications to the base model. The same prediction head is used to generate the final prediction based on the aggregated features.

\subsection{Fusion Aggregation Module} \label{subsec:FAM}

\begin{figure}
    \centering
    \includegraphics[width=1.0\linewidth]{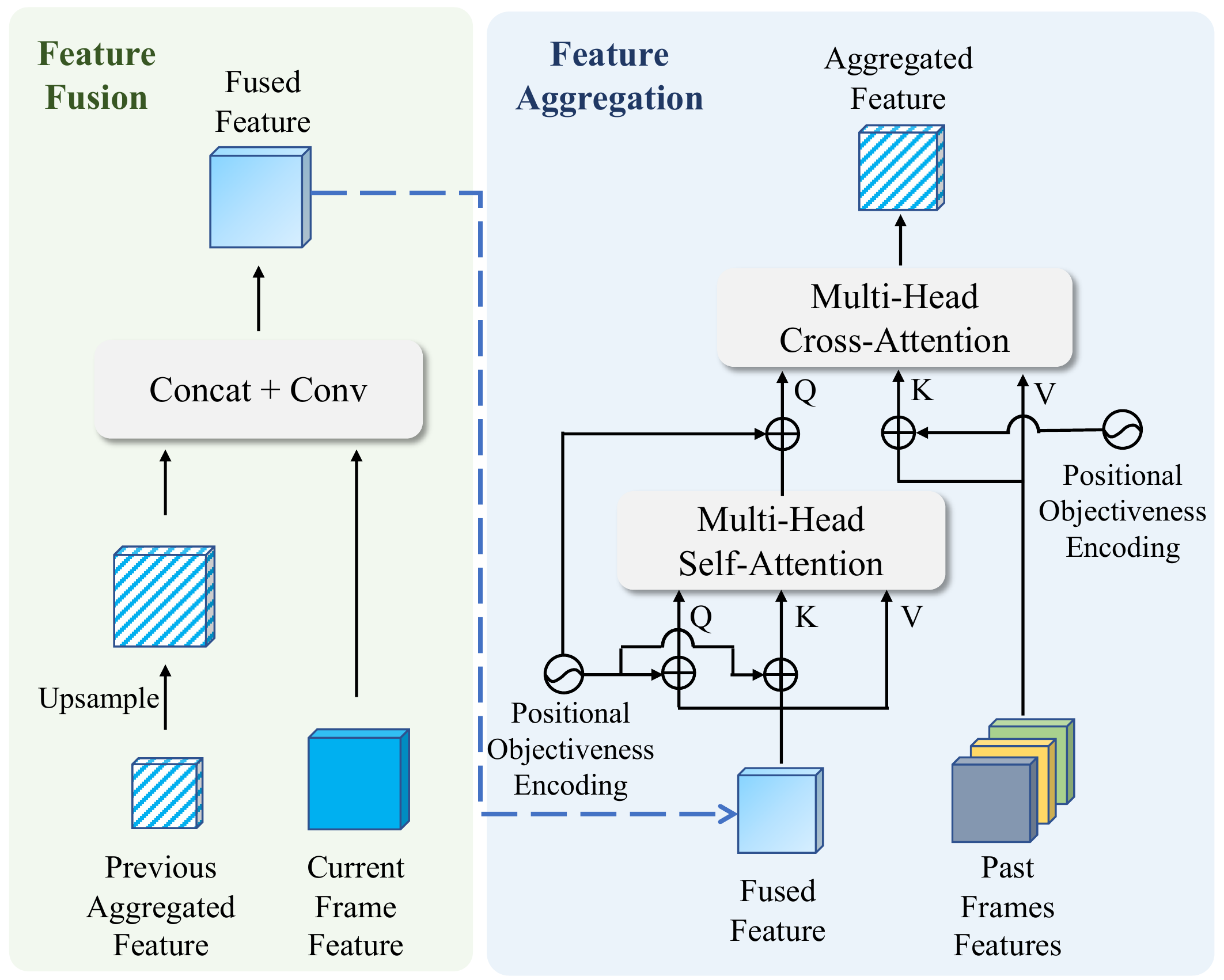}
    \caption{
    Architecture of the proposed Fusion Aggregation Module (FAM): FAM fuses (optional) and aggregates features of each scale level. For feature fusion, the previously aggregated feature is first upsampled and then fused with the current frame feature. For feature aggregation, the fused feature is aggregated with features of past frames using a transformer. 
    }
\label{fig:fusion_aggregation}
\vspace{-2mm}
\end{figure}

Fig.~\ref{fig:fusion_aggregation} illustrates the architecture of the Fusion Aggregation Module (FAM). The FAM consists of feature fusion and feature aggregation operations. The feature fusion is an operation that takes as input the aggregated feature $\hat{\mathbf{F}}^{i-1}_{T}$ from the previous FAM as well as the current frame feature ${\mathbf{F}}_{T}^{i}$, where $i$ denotes the scale level. The previous aggregated feature $\hat{\mathbf{F}}^{i-1}_{T}$ is upsampled by a transpose convolution layer to match the size of the current scale\, level\, and\, concatenated with the current frame feature ${\mathbf{F}}_{T}^{i}$. A convolution layer is used to fuse the two concatenated inputs and the fused feature $\Bar{\mathbf{F}}_{T}^{i}$ is obtained by:
\begin{align}
    \Bar{\mathbf{F}}_{T}^{i} = \text{Conv}([\text{upsample}(\hat{\mathbf{F}}^{i-1}_{T}), \mathbf{F}_T^i])
\end{align}
where $[\cdot]$ denotes concatenation. By fusing the previous aggregated feature of smaller scale with the current frame feature, the fused feature integrates the coarse cross-frame matching information, which is used to guide the subsequent feature aggregation. Note that feature fusion is not required for the smallest scale level since it is the first level.

In the feature aggregation stage, the fused feature $\Bar{\mathbf{F}}_{T}^{i}$ aggregates information from the features from past frames denoted by $\{\mathbf{F}_{T-n}^i\}_{n=1}^{N-1}$, where $N$ is the number of frames. As illustrated in Fig.~\ref{fig:fusion_aggregation}, the fused feature is first processed by a multi-head self-attention module to gather features within the frame, and the output feature serves as the query for the subsequent cross-attention module which performs cross-frame feature aggregation, while the features of the past frames are concatenated and serve as the key and value. The feature aggregation process can be summarized as:
\begin{align}
    \hat{\mathbf{F}}_T^i = \text{Attn}(\text{Attn}(\Bar{\mathbf{F}}_T^i, \Bar{\mathbf{F}}_T^i), [\mathbf{F}_{T-1}^i, ..., \mathbf{F}_{T-N+1}^i])
\end{align}
where the output $\hat{\mathbf{F}}_T^i$ is the aggregated feature of the current scale $i$. Note that the normalization layer, skip connection and feed-forward network in the transformer are omitted for simplicity, and each FAM consists of $L$ Transformer layers described above for feature aggregation.

\noindent\textbf{Positional Objectiveness Encoding.\;\;\;} 
On top of the regular positional encoding generated with sinusoidal functions as introduced in \cite{vaswani2017attention}, we incorporate an additional objectiveness encoding to facilitate the feature aggregation process. Specifically, we obtain the classification prediction from the base model for each frame and use a convolution layer to encode it into the same dimension as the feature maps. In the case of multi-class prediction, the highest score is selected. Formally, the objectiveness encoding $E_{obj}$ is obtained by:
\begin{align}
    E_{obj} = \text{Conv}(\sigma(\max_{c=1}^C{S}))
\end{align}
where $C$ denotes the number of classes and $S$ represents the classification prediction. $\sigma(\cdot)$ is the sigmoid function that converts the scores to the [0,1] range. The objectiveness encoding is then summed with the positional encoding to form the positional objectiveness encoding. The classification prediction reflects the objectiveness of each token in the feature map, and explicitly adding this information to the query and key features helps the transformer locate foreground objects during the matching process.

\begin{figure}[t]
    \centering
    \includegraphics[width=1.0\linewidth]{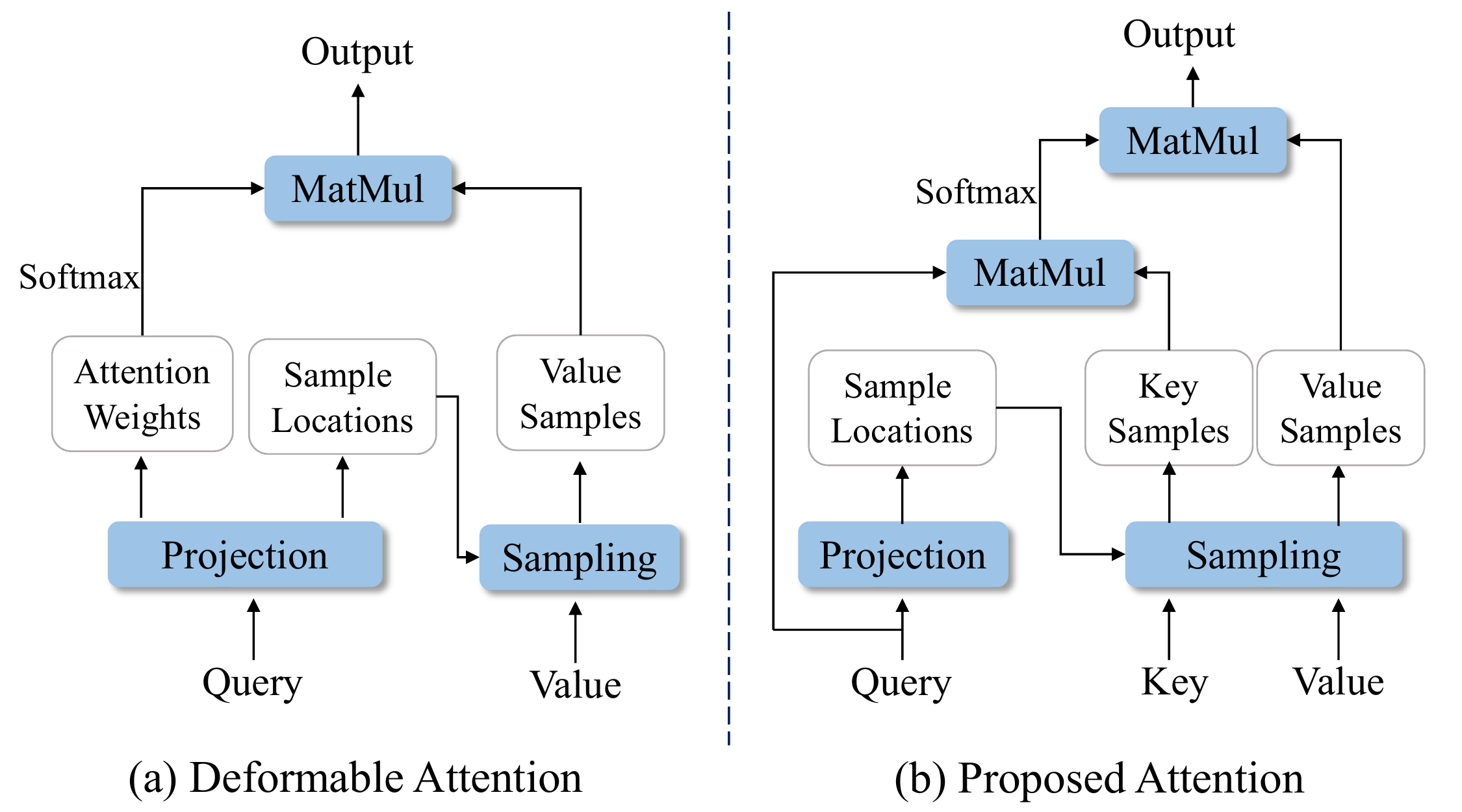}
    \caption{Comparison of the original deformable attention \cite{zhu2020deformable} and our proposed variant. (a) The original deformable attention generates the attention weights directly via projection from the query features. (b) Our proposed attention samples key elements and performs query-key matching to obtain the attention matrix, which is applied on the value samples to yield the output.
    }
    \label{fig:attention}
    \vspace{-2mm}
\end{figure}

\subsection{Attention Mechanism} \label{subsec:attention}

One known issue of regular transformers is their high computational complexity and memory consumption since global matching is performed between query and key tokens. The complexity grows quadratically as the number of tokens increases. This issue is even amplified as we deal with large-scale point cloud scenes. Deformable attention \cite{zhu2020deformable} is proposed in Deformable DETR as an alternative attention mechanism to reduce the computation and speed up the convergence of the detection model. As shown in Fig.\ref{fig:attention}(a), instead of computing the global attention matrix, deformable attention generates a small number of sampling locations based on the position of the query element through a linear projection from the query feature. At the same time, the query feature is also projected to a set of attention weights corresponding to the sampling locations. Value samples are obtained from the value feature based on the sampling locations, and the output is computed by multiplying the attention weights with the value samples. 

Deformable attention greatly reduces the computation complexity from quad-ratic to linear w.r.t. the feature length. Moreover, it does not restrict the attention pattern to a fixed local range, which makes it favorable for our application where objects are of different moving speeds. However, the original deformable attention does not explicitly enforce query-key matching for attention matrix generation, which limits the capability in modeling the cross-frame correlation of moving objects in our task. 
Specifically, when the current frame feature serves as the query in cross-frame matching, the value feature comes from the past frames and might be misaligned with the query due to object movements. However, the attention weights are directly generated by projecting the query feature, which does not contain the motion information. Therefore, it is difficult for the attention module to generate meaningful attention weights in order to focus on the moving objects (as shown in Fig.\ref{fig:attention_vis}). 
To address this issue, we introduce a variant of deformable attention by incorporating the query-key matching process. Concretely, the projected sampling locations are used to generate key samples and value samples, respectively. The key samples are multiplied with the query feature to obtain the attention matrix, which is further applied to the value samples to generate the final output. Formally, we use $\{\mathbf{q}_i\}$ to denote a set of query tokens as well as $\{\mathbf{k}_i\}$ and $\{\mathbf{v}_i\}$ for sampled key and value tokens. The attention matrix is computed by:
\begin{align}
    A_{i,j}^h = \text{softmax}(\frac{(W_q \mathbf{q}_i )^T(W_k \mathbf{k}_j )} {\sqrt{d}})
\end{align}
where h indexes the attention heads, and $W_q$ and $W_k$ are the query and key projections. $\sqrt{d}$ is a scaling factor \cite{vaswani2017attention} while $d$ is the feature dimension. The output of the attention is calculated by:
\begin{align}
    \text{Attn}(\mathbf{q}_i, \mathbf{k}_j, \mathbf{v}_j) = \sum_{h=1}^H{W_o}(\sum_{j=1}^K{A_{i,j}^h}\cdot W_v \mathbf{v}_j)
\end{align}
where $W_o$ and $W_v$ are the output projection and value projection, respectively. $K$ denotes the number of sample locations for each query element and $H$ is the number of heads. 

\begin{table*}[t]
\begin{center}
\caption{Performance comparison with multi-frame 3D detectors on the nuScenes dataset. T.C., Moto. and Cons. stand for traffic cone, motorcycle, and construction vehicle, respectively. Mean Average Precision (mAP) is used for evaluation.}
\vspace{-2mm}
\label{tab:nuscenes}
\setlength{\tabcolsep}{4pt}
\scalebox{0.9}{
\begin{tabular}{c|ccccccccccc}
\toprule[1.25pt]
Method & Car & Ped & Bus & Barrier & T.C. & Truck & Trailer & Moto. & Cons. & Bicycle & Mean \\ \midrule[0.85pt]
3DVID \cite{yin2020lidarconvgru}  & 79.7 & 76.5 & 47.1 & 48.8 & \textbf{58.8} & 33.6 & \textbf{43.0} & 40.7 & 18.1 & 7.9 & 45.4 \\
TCTR \cite{yuan2021temporalTCTR} & 83.2 & 74.9 & \textbf{63.7} & 53.8 & 52.5 & 51.5 & 33.0 & 54.0 & 15.6 & 22.6 & 50.5 \\
Ours  & \textbf{84.0} & \textbf{77.9} & 62.0 & \textbf{55.1} & 55.4 & \textbf{52.4} & 34.3 & \textbf{55.2} & \textbf{18.9} & \textbf{27.6} & \textbf{52.3} \\
\bottomrule[1.25pt]
\end {tabular}
}
\end{center}
\vspace{-3mm}
\end{table*}

\subsection{Losses}
\label{subsec:losses}
In the feature aggregation stage of FAMs, apart from the final aggregated feature, the intermediate outputs from each transformer layer is also kept to provide additional supervision. We fuse the multi-scale features in the same manner as the base model \cite{lang2019pointpillars} to generate the final prediction for loss computation. The losses are the same as the base model, which consist of the classification loss $\mathcal{L}_{cls}$, the localization loss $\mathcal{L}_{loc}$, and the direction loss $\mathcal{L}_{dir}$. We sum the losses for all the transformer layers to get the aggregation loss:
\begin{align}
    \mathcal{L}_{aggr} = \frac{1}{L} \sum_{l=1}^L (\beta_{cls}\mathcal{L}_{cls} + \beta_{loc}\mathcal{L}_{loc} +\beta_{dir}\mathcal{L}_{dir})
\end{align}
where $L$ is the number of transformer layers and $\beta_{cls}$, $\beta_{loc}$, $\beta_{dir}$ are the loss coefficients. On the other hand, the base model loss is computed for all the input frames, and the aggregation loss and the base model loss are summed to get the total loss:
\begin{align}
    \mathcal{L} = \mathcal{L}_{base} + \mathcal{L}_{aggr}
\end{align}
The whole model is optimized in an end-to-end manner.

\section{Experiments}

\subsection{Datasets}
We evaluate our proposed method on two popular point cloud detection benchmarks that include sequential frames.

\noindent\textbf{nuScenes.\;\;\;}
The nuScenes \cite{caesar2020nuscenes} dataset contains 700 sequences for training and 150 for validation. Each point cloud sequence is around 20s in length with a frame interval of 0.05s. Annotations are provided for every 10 consecutive frames, which are known as key frames. The main evaluation metric for the detection task is mean average precision (mAP). The mAP calculation uses a series of center distance thresholds instead of the commonly used box IoU threshold. 

\noindent\textbf{Waymo.\;\;\;}
The Waymo Open Dataset \cite{sun2020waymo} is a large-scale autonomous driving dataset that contains 798 point cloud sequences for training and 202 sequences for validation. The point clouds are collected using 64-line LiDAR, which gives around 180k points for each frame. The frame interval between consecutive frames is 0.1s and the annotations are provided for each frame in the training data. The detection performance is evaluated by mean average precision (mAP) and mAP weighted by heading accuracy (mAPH) while the objects are categories into 2 difficulty levels based on the number of points contained.

\subsection{Implementation Details}
\noindent\textbf{nuScenes.\;\;\;}
We use an input range of [-51.2m, 51.2m] for the x-y plane, and [-5m, 3m] for the z-axis. The voxel size is set to 0.2m, which leads to $512\times512$ pillars in total. The initial feature maps are downsampled with factors of [2, 4, 8] to obtain the multi-scale feature maps as described in \cite{lang2019pointpillars}. Ego-motion correction is performed to compensate the self-motion. 
We concatenate the points from the associated sweeps with the key frame to form one input frame. We follow the practice in \cite{yin2020lidarconvgru, yuan2021temporalTCTR} to use 3 key frames and their associated sweeps as input. For the FAM, we use 6 transformer layers with 8 attention heads. The number of deformable points is set as 8. We select voxels with top $\text{5\%}$ classification scores to form the query feature. The base model is trained for 40 epochs with a learning rate of 0.001 and cosine annealing scheduling, and the whole model is then trained for 30 epochs with a learning rate 0.0005. 

\noindent\textbf{Waymo.\;\;\;}
For the experiments on Waymo, we use input point cloud range of [-76.8m, 76.8m] for x- and y-axis and [-2m, 4m] for the z-axis, following the settings in \cite{yang20213dman}. The voxel size is set as 0.3m, which gives us the same number of pillars as the nuScenes dataset.
We follow \cite{yang20213dman} to use 16 frames as the input by dividing them into 4 windows of 4 frames, where the points inside each window are concatenated. 
The base model is first trained for 20 epochs with a learning rate of 0.003, and we then train the whole model for another 10 epochs with a learning rate of 0.0016. The rest of the settings are the same as nuScenes.

\begin{table*}[t]
\begin{center}
\setlength{\tabcolsep}{4pt}
\caption{Performance comparison with multi-frame 3D detectors on the Waymo validation dataset. Even with a less competitive baseline model, our method outperforms the state-of-the-art multi-frame detector 3D-MAN. The difference regarding baseline performance is attributed to the modifications introduced by 3D-MAN, which is not open-sourced.}
\label{tab:waymo}
\vspace{-2mm}
\scalebox{0.9}{
\begin{tabular}{ c|c| cccc | cccc}
\toprule[1.25pt]
\multirow{2}{*}{Difficulty} & \multirow{2}{*}{Method} & \multicolumn{4}{ c |}{mAP (IoU=0.7)} & \multicolumn{4}{ c }{mAPH (IoU=0.7)}  \\ 
\cline{3-10}
&  & Overall  & 0-30m & 30-50m & 50m-Inf & Overall  & 0-30m & 30-50m & 50m-Inf \\ \midrule[0.85pt]
\multirow{5}{*}{Level 1} 
    & ConvLSTM \cite{huang2020lstm} & 63.6  &- & - & - & - & - & - & -   \\ \cline{2-10}
    & 3D-MAN\,(baseline) & 69.03 & 87.99 & 66.55 & 43.15 & 68.52 & 87.57 & 65.92 & 42.37 \\
    & 3D-MAN \cite{yang20213dman} & 74.53 & \textbf{92.19} & 72.77 & 51.66 & 74.03 & \textbf{91.76} & 72.15 & 51.02  \\ \cline{2-10}
    & Ours\,(baseline) & 67.40 & 86.91 & 62.93 & 41.35 & 66.74 & 86.34 & 62.19 & 40.47 \\ 
    & Ours & \textbf{74.97} & 91.39 & \textbf{73.10} & \textbf{53.52} & \textbf{74.42} & 90.91 & \textbf{72.48} & \textbf{52.75} \\ \midrule[0.85pt]

\multirow{4}{*}{Level 2} 
     & 3D-MAN\,(baseline) & 60.16 & 87.10 & 59.27 & 32.69 & 59.71 & 86.68 & 58.71 & 32.08  \\ 
     & 3D-MAN \cite{yang20213dman} & 67.61 & \textbf{92.00} & 67.20 & 41.38 & 67.14 & \textbf{91.57} & 66.62 & 40.84  \\ \cline{2-10}
     & Ours\,(baseline) & 58.94 & 86.07 & 57.13 & 31.72 & 58.36 & 85.50 & 56.44 & 31.02 \\
     & Ours & \textbf{67.89} & 90.69 & \textbf{67.37} & \textbf{42.57} & \textbf{67.35} & 90.20 & \textbf{66.78} & \textbf{41.91} \\
\bottomrule[1.25pt]
\end{tabular}
}
\vspace{-4.0mm}
\end{center}
\end{table*}

\subsection{Benchmarking Results}

\begin{figure}[t]
    \centering
    \vspace{-3mm}
    \subfloat{{\includegraphics[width=4.2cm]{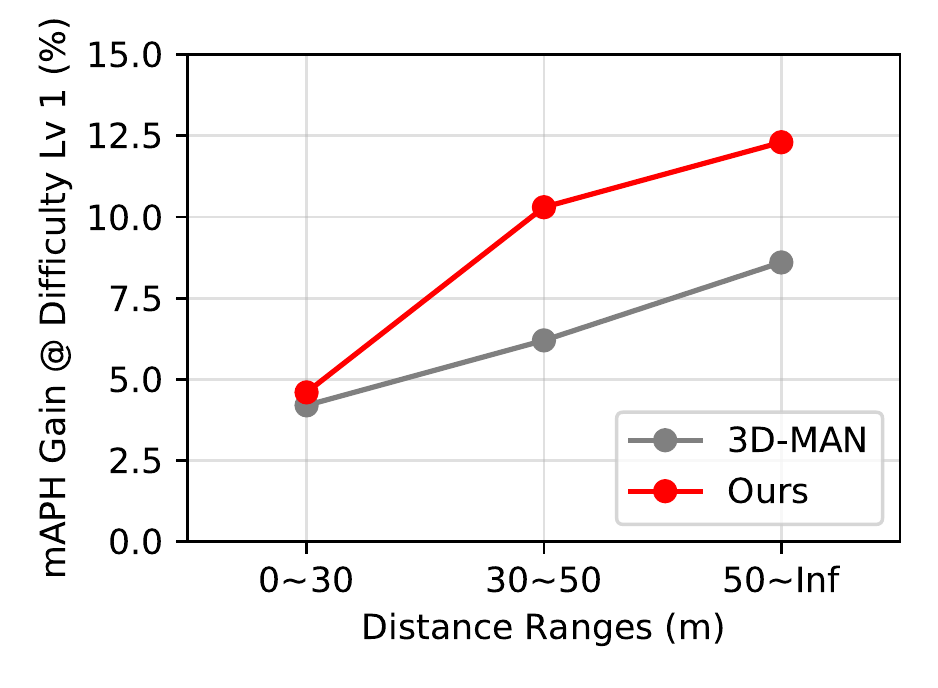} }}%
    \subfloat{{\includegraphics[width=4.2cm]{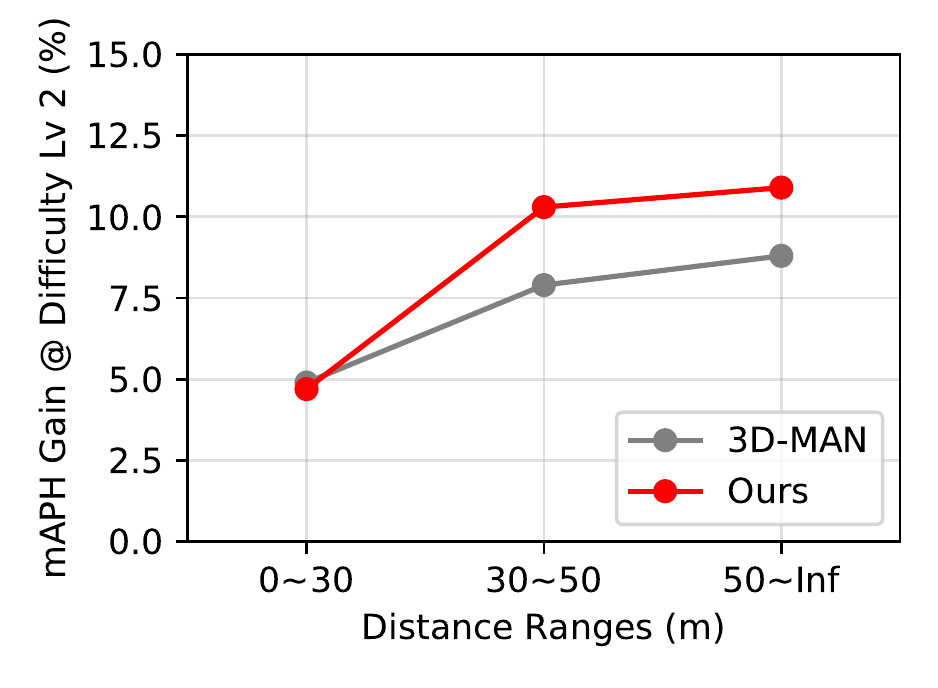} }}%
    \vspace{-4mm}
    \caption{Compared with 3D-MAN, our proposed method has a more significant performance gain for objects that are further away ($>$30m), which are more susceptible to point sparsity and occlusions, thus benefiting more from multi-frame information.}%
    \vspace{-3mm}
    \label{fig:maph_gain_comparison}%
\end{figure}

We compare with existing multi-frame detection methods that exploit temporal information to improve the detection performance. We do not include single-frame detectors as they build on distinct architectures and parameters, which are not the focus of this study.
Table~\ref{tab:nuscenes} reports the detection performance on the nuScenes dataset. 
Our proposed method outperforms the current start-of-the-art multi-frame detection method TCTR \cite{yuan2021temporalTCTR} by a margin of 1.8 in mAP while achieving the best result for the majority of the categories. Table~\ref{tab:waymo} reports the detection performance on the Waymo dataset. \modelname{} outperforms the state-of-the-art multi-frame method 3D-MAN \cite{yang20213dman}: although 3D-MAN employs a stronger base model than ours, we still achieve better results with convincingly larger margins of improvements upon the base models. The difference in the base model performance comes from the modifications made to the backbone by 3D-MAN\footnote{3D-MAN has yet open-sourced its implementation.}, while we use the original PointPillars \cite{lang2019pointpillars}. From the distance breakdown, it is observed from Fig.~\ref{fig:maph_gain_comparison} that our model achieves higher performance gain for objects at a longer distance ($>$30m) compared to 3D-MAN, e.g. an improvement of 10.29 in mAPH for level 1 objects in the 30-50m range as compared to the 6.23 increment for 3D-MAN. As pointed out in \cite{yang20213dman}, further objects typically have sparser point distributions and are more susceptible to occlusions, which make the multi-frame information more useful in complementing the single-frame view. The significant performance gain demonstrates the effectiveness of our method in aggregating multi-frame features.

\begin{table}[t]
\begin{center}
\caption{Ablation study on different aggregation methods. 
}
\label{tab:aggregation} 
\vspace{-2mm}
\setlength{\tabcolsep}{4.0pt}
\scalebox{0.8}{
\begin{tabular}{c|cc|cc}
\toprule[1.25pt]
\multirow{2}{*}{Aggregation Method} & \multicolumn{2}{ c|}{Level 1} & \multicolumn{2}{c}{Level 2} \\ \cline{2-5}
&  mAP & mAPH & mAP & mAPH  \\
\midrule[0.85pt]
Single-Scale & 59.78 & 58.86 & 52.05 & 51.26 \\
Separate Multi-Scale & 61.21 & 60.44 & 53.38 & 52.70 \\
Hierarchical Multi-Scale (Ours) & 62.09 & 61.23 & 55.45 & 54.59 \\
\bottomrule[1.25pt]
\end {tabular}
}
\end{center}
\vspace{-5.0mm}
\end {table}

\subsection{Ablation Studies}

We conduct ablation studies to investigate the effectiveness of our proposed components on the Waymo dataset. Due to the overwhelming number of training samples and limited computational resources, we sample 10\% of sequences from the training set for the experiments. We use 4 point cloud frames with an interval of 0.4s between adjacent frames as the input to report the results unless specified otherwise. The evaluation is conducted with the full validation dataset.

\noindent\textbf{Aggregation Methods.\;\;\;}
We compare our proposed hierarchical coarse-to-fine feature aggregation method with two different approaches, namely \textit{Single-scale} and \textit{Separate Multi-scale}. For the Single-scale setting, the final concatenated feature maps of the base model are used for feature aggregation. Separate Multi-scale refers to performing feature aggregation for different scales in parallel and then combining the aggregated features. As shown in Table \ref{tab:aggregation}, we observe that the performance drops for both methods while Single-scale suffers a larger loss. It shows that multi-scale features are helpful in learning cross-frame relations and our proposed hierarchical coarse-to-fine aggregation strategy makes the feature aggregation more effective.

\noindent\textbf{Attention Mechanisms.\;\;\;}
We compare our proposed attention mechanism with the original deformable attention \cite{zhu2020deformable} and report the results in Table \ref{tab:ablation_attention}. The model using the original deformable attention \cite{zhu2020deformable} experiences a clear performance drop as compared to our proposed attention mechanism, which shows the importance of query-key matching for aggregating multi-frame features in the task this paper studies.

\noindent\textbf{Transformer Encoding.\;\;\;}
To investigate the effectiveness of our proposed positional objectiveness encoding, we conduct experiments by removing the positional encoding and the objectiveness encoding. The results in Table \ref{tab:ablation_embedding} show that the removal of each type of encoding leads to a loss in model performance, while the objectiveness encoding has a more significant impact. It demonstrates that the objectiveness encoding has a positive effect on guiding the feature aggregation by adding an additional signal to indicate the objectiveness of each input token.

\begin{table}[t]
    \centering
\caption{Ablation study on attention mechanisms.}
\vspace{-2mm}
\label{tab:ablation_attention} 
      \setlength{\tabcolsep}{2.5pt}
         \scalebox{0.9}{
         \begin{tabular}{c|cc|cc}
\toprule[1.1pt]
\multirow{2}{*}{Attention Type} & \multicolumn{2}{ c|}{Level 1} & \multicolumn{2}{c}{Level 2} \\ \cline{2-5}
&  mAP & mAPH & mAP & mAPH  \\
\midrule[0.75pt]
DeformAttention\,\cite{zhu2020deformable} & 60.86 & 60.04 & 53.03 & 52.32 \\
Ours & 62.09 & 61.23 & 55.45 & 54.59\\
\bottomrule[1.1pt]
\end {tabular}
}
\end{table}

\begin{table}[t]
    \centering
\caption{Ablation study on transformer encodings.}
\vspace{-2mm}
\label{tab:ablation_embedding} 
\setlength{\tabcolsep}{2.75pt}
\scalebox{0.9}{
\begin{tabular}{c|cc|cc}
\toprule[1.25pt]
\multirow{2}{*}{Encoding Type} & \multicolumn{2}{ c|}{Level 1} & \multicolumn{2}{c}{Level 2} \\ \cline{2-5}
 &  mAP & mAPH & mAP & mAPH  \\
\midrule[0.85pt]
w/o Pos. Encoding & 61.52	& 60.68	& 54.87	& 54.09 \\
w/o Obj. Encoding & 61.29	& 60.47	& 53.32	& 52.60\\
Ours & 62.09 & 61.23 & 55.45 & 54.59 \\
\bottomrule[1.25pt]
\end {tabular}}
\vspace{-2mm}
\end{table}

\subsection{Further Discussions}

\noindent\textbf{Number of Input Frames.\;\;\;} We study the performance of our proposed method under different numbers of input frames. For the experiments with 8 and 16 input frames, we divide them into 4 windows of 2 and 4 concatenated frames, respectively. Additionally, we also compare with the basic point concatenation~\cite{caesar2020nuscenes} method using 4 frames (denoted as $4^*$). As shown in Table~\ref{tab:frames}, by using 4 consecutive frames as input, there is already a substantial improvement over the single-frame baseline (e.g. an increase of 6.86 in mAP for level 1 objects.). Our method outperforms the point concatenation method by a clear margin when using the same number of frames, which demonstrates the benefit of explicit cross-frame correlation modeling. As the number of frames is further increased, we notice a higher improvement for level 2 objects, which is in line with the observation that objects with sparse point distributions can benefit more from multi-frame information.

\begin{table}[]
    \centering
    \vspace{-0.0mm}
    \caption{Performance comparison on number of input frames.\; $*$ denotes point concatenation.}
    \label{tab:frames} 
    \vspace{-1.0mm}
    \setlength{\tabcolsep}{3.5pt}
    \scalebox{0.9}{
    \begin{tabular}{c|cc|cc}
    \toprule[1.25pt]
    Number of & \multicolumn{2}{ c|}{Level 1} & \multicolumn{2}{c}{Level 2} \\ \cline{2-5}
    Frames &  mAP & mAPH & mAP & mAPH  \\
    \midrule[0.85pt]
    1  & 54.43	& 53.52	& 47.04	& 46.24 \\
    $\text{\space4}^*$ & 57.01 & 56.21	& 50.39	& 49.66 \\ 
    4 & 60.75 & 59.91 & 52.76 & 52.02 \\
    8 & 61.26 & 60.43 & 53.36 & 52.64 \\
    16 & 62.16	& 61.41	& 55.55	& 54.85 \\
    \bottomrule[1.25pt]
    \end {tabular}
    }
    \vspace{-3.00mm}
\end{table}

\begin{figure}
    \centering
    \includegraphics[width=1.0\linewidth]{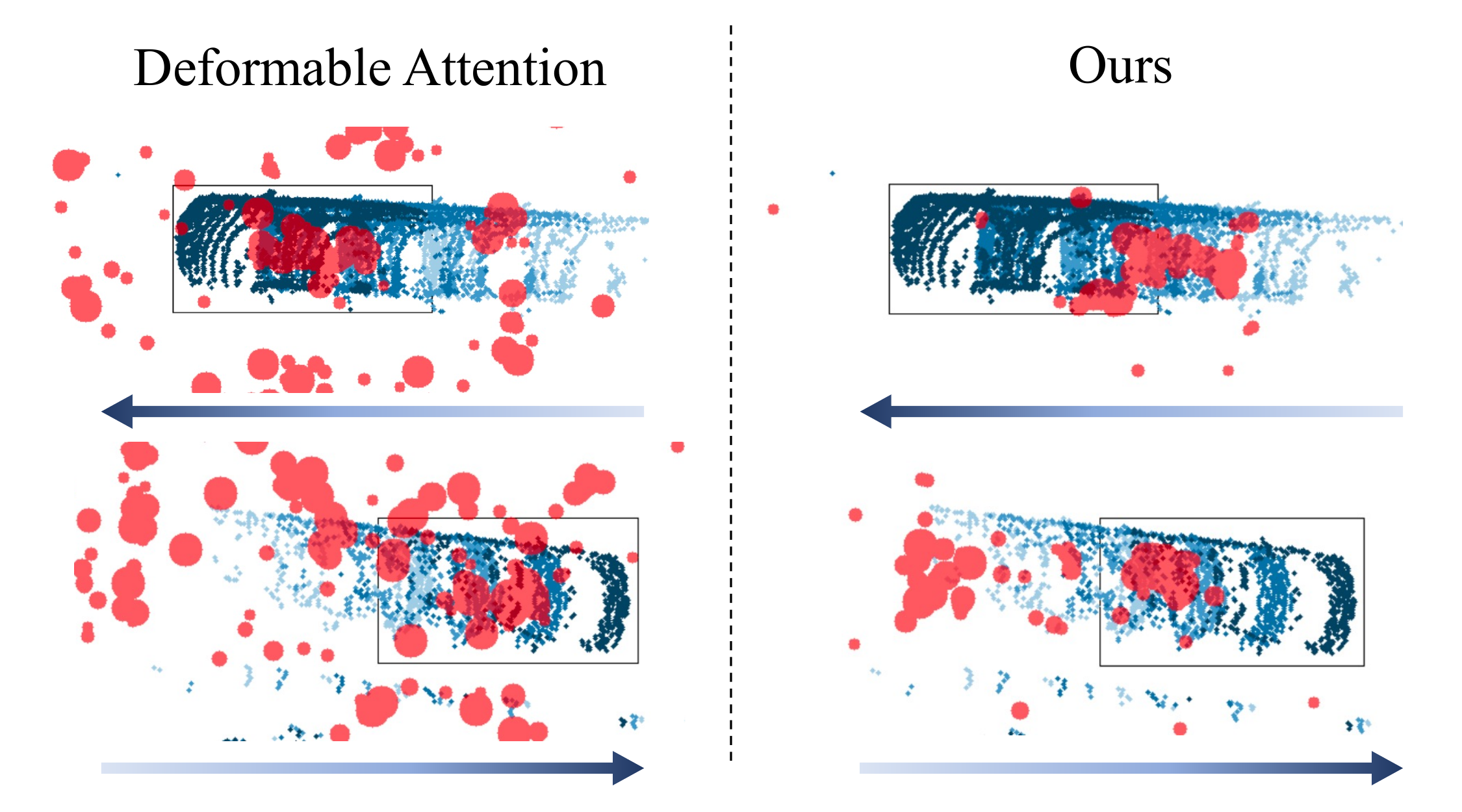}
    \vspace{-2mm}
    \caption{
    Visualization of attention weights. Current frame points are shown in dark blue, and the intensity decreases as the time interval increases for past frame points. The arrows indicate the objects' moving directions. The attention sampling locations are shown in red, while the radius of each circle indicates the attention weight magnitude.
    }
    \label{fig:attention_vis}
    \vspace{-2mm}
\end{figure}

\noindent\textbf{Attention Visualization.\;\;\;}
We visualize the learned attention patterns on the bird eye's view of object instances. As illustrated in Fig.~\ref{fig:attention_vis}, we use red circles to represent the attention sampling locations while the radius of each circle indicates the attention magnitude. It is observed that the original deformable attention \cite{zhu2020deformable} demonstrates highly sparse attention patterns haphazardly scattered around the object, while our proposed attention mechanism effectively focuses on the trajectories of moving objects. It demonstrates that the proposed query-key matching operation is essential to the cross-frame feature aggregation task as it is more aware of the object motion in the aggregation process.

\section{Conclusion}
In this paper, we present a multi-frame 3D object detection method named \modelname{}, which performs voxel-level feature aggregation. We design a novel hierarchical aggregation strategy to perform coarse-to-fine aggregation based on multi-scale features: coarse features are used to capture object motions to guide the fusion of fine features for accurate localization. A variant of deformable attention is introduced for efficient and motion-aware feature matching. Experimental results show that \modelname{} outperforms state-of-the-art multi-frame detection methods on standard benchmarks.

{\small
\bibliographystyle{ieee_fullname}
\bibliography{egbib}
}

\end{document}